\documentclass[sigconf,nonacm]{acmart}

\usepackage{ulem}
\usepackage[shortlabels]{enumitem} 
\usepackage{graphicx}
\usepackage{amsmath}
\usepackage{mathtools}
\usepackage{comment}
\usepackage{bm}
\usepackage{multirow}
\usepackage{threeparttable,booktabs}
\usepackage{tikz}
\usepackage{balance}
\usepackage{courier}                                       
\usepackage{cleveref}                                      
\usepackage[mathcal]{eucal}
\usepackage[noend]{algpseudocode}
\usepackage[ruled,lined,linesnumbered,noend]{algorithm2e}
\usepackage{pgfplots}
\usepackage{pgfplotstable}
\pgfplotsset{compat=newest}
\usepackage{siunitx}
\theoremstyle{plain}

\theoremstyle{definition}

\usepackage[skip=1pt]{caption}            
\usepackage{graphicx}
\PassOptionsToPackage{table}{xcolor}
\usepackage{subcaption}
\usepackage{stfloats}
\usepackage{multirow}
\usepackage{booktabs}
\usepackage{colortbl}
\usepackage{rotating}
\usepackage{tikz}

\graphicspath{{./figs/}{./doc/figs/}}

\definecolor{green(html/cssgreen)}{rgb}{0.0, 0.5, 0.0}

\setcopyright{none}
\copyrightyear{2026}
\acmYear{2026}
\acmConference[ICCAD '26]{IEEE/ACM International Conference on Computer-Aided Design}{November 8--12, 2026}{San Jose, CA, USA}
\acmBooktitle{Proceedings of the IEEE/ACM International Conference on Computer-Aided Design (ICCAD '26), November 8--12, 2026, San Jose, CA, USA}
\acmDOI{}
\acmISBN{}

\begin{document}

\title{\texorpdfstring{\texttt{GoGoTB}}{GoGoTB}: Agentic RTL Verification with Specification-Grounded Coverage Closure}

\makeatletter
\def\@authorfont{\large}
\def\@affiliationfont{\small}
\makeatother

\newcommand{\gogotbauthorsep}{,\enspace}
\newcommand{\gogotbfirstequal}{1,\textdagger}
\newcommand{\gogotbsecondequal}{2,\textdagger}
\newcommand{\gogotbyibomultiaffiliation}{2,4,5}
\newcommand{\gogotbcofirstnote}{%
  Xin Xin and Jincheng Lou contributed equally to this work.}

\author[Xin et al.]{%
  Xin Xin\textsuperscript{\gogotbfirstequal}\gogotbauthorsep
  Jincheng Lou\textsuperscript{\gogotbsecondequal}\gogotbauthorsep
  Junhui Li\textsuperscript{3}\gogotbauthorsep
  Jinglin Yan\textsuperscript{1}\gogotbauthorsep
  Panda Xiao\textsuperscript{1}\gogotbauthorsep
  Di Wu\textsuperscript{1}\gogotbauthorsep
  Haixiao Li\textsuperscript{1}\\[0.15em]
  Weicong Lu\textsuperscript{1}\gogotbauthorsep
  Weijian Fan\textsuperscript{2}\gogotbauthorsep
  Xinyu Qu\textsuperscript{2}\gogotbauthorsep
  Yuxiang Zhao\textsuperscript{2}\gogotbauthorsep
  Min Yu\textsuperscript{2}\gogotbauthorsep
  Zhixiong Di\textsuperscript{3}\gogotbauthorsep
  Yibo Lin\textsuperscript{\gogotbyibomultiaffiliation}%
}

\affiliation{%
  \institution{%
    \textsuperscript{1}Tencent\enspace
    \textsuperscript{2}School of Integrated Circuits,
    Peking University, Beijing\\
    \textsuperscript{3}Southwest Jiaotong University\enspace
    \textsuperscript{4}Institute of Electronic Design Automation,
    Peking University, Wuxi\\
    \textsuperscript{5}Beijing Advanced Innovation Center for
    Integrated Circuits\\[0.25em]
    \scriptsize
    \texttt{\{wukongxin,lyndayan,pandaxiao,woodywdwu,stevehxli,lukewclu\}@tencent.com}\\
    \texttt{\{jinchenglou,wjfan25,xyqu25,yuxiangzhao\}@stu.pku.edu.cn}\\
    \texttt{ssfortynine@gmail.com}\enspace
    \texttt{yum@pku.edu.cn}\enspace
    \texttt{dizhixiong2@126.com}\enspace
    \texttt{yibolin@pku.edu.cn}
  }
  \city{}
  \country{}
}

\renewcommand{\authors}{Xin Xin\and Jincheng Lou\and Junhui Li\and
  Jinglin Yan\and Panda Xiao\and Di Wu\and Haixiao Li\and Weicong Lu\and
  Weijian Fan\and Xinyu Qu\and Yuxiang Zhao\and Min Yu\and Zhixiong Di\and
  Yibo Lin}
\renewcommand{\shortauthors}{Xin et al.}



\begin{abstract}
Functional verification dominates integrated circuit (IC) front-end engineering effort, and a single missed bug that escapes to silicon can trigger a costly respin.
Recent large language models (LLMs) offer new opportunities to automate this process, yet existing LLM-based approaches generate each component through independent single-turn calls with no shared context, leaving interface mismatches undetected and reported coverage disconnected from specification requirements.
To address these challenges, we present \texttt{GoGoTB}, an agentic framework that achieves end-to-end verification closure through three subsystems: an agentic execution control layer, an evolvable knowledge system, and specification-grounded coverage closure.
The execution control layer separates deterministic enforcement from LLM reasoning at every tool and stage boundary.
The knowledge system dispatches methodology and design-specific expertise on demand.
The coverage framework anchors every bin to a named specification behavior so that each residual gap has a diagnosable root cause and a targeted remedy.
Tested on 8 register transfer level (RTL) designs without any human intervention, \texttt{GoGoTB} achieves 100\% environment generation success and averages 98.4\% line, 97.2\% branch, 97.0\% toggle, and 83.2\% functional coverage.
No prior work successfully generates a complete verification environment or achieves meaningful coverage on the same benchmarks.
\end{abstract}

\maketitle
\begingroup
\renewcommand{\thefootnote}{\textdagger}
\footnotetext{\gogotbcofirstnote}
\endgroup
\newcommand\blfootnote[1]{%
  \begingroup
  \renewcommand\thefootnote{}\footnote{#1}%
  \addtocounter{footnote}{-1}%
  \endgroup
}

\section{Introduction}
\label{sec:intro}

\begin{figure}[t]
  \centering
  \includegraphics[width=\columnwidth]{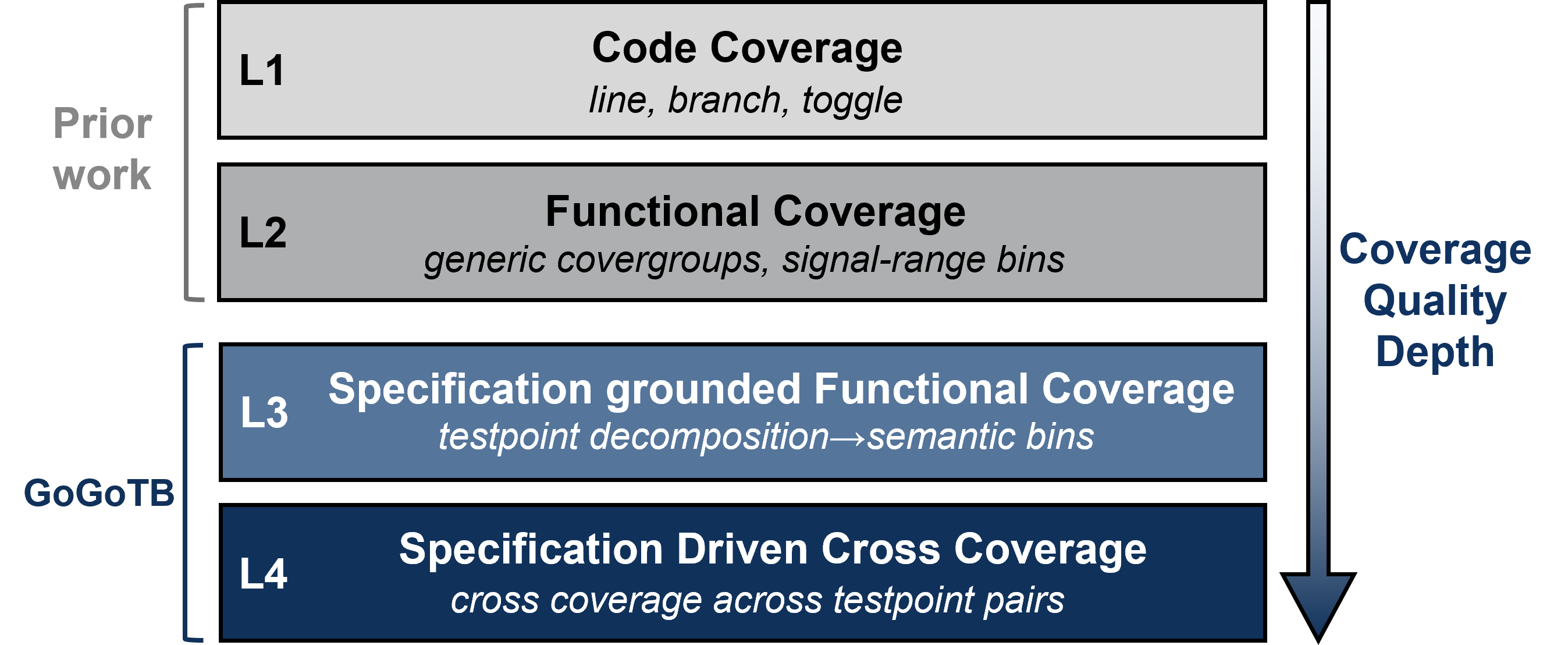}
  \caption{Four-level verification coverage hierarchy. Prior work covers L1--L2. \texttt{GoGoTB} delivers L3--L4.}
  \label{fig:teaser}
\end{figure}

Functional verification consumes the majority of front-end engineering effort in IC development~\cite{IEEEUVM2017,Foster2015Trends}, and assembling a complete UVM environment demands weeks of specialized effort per design~\cite{Bergeron2003Writing,Spear2008SystemVerilog}. Recent LLMs have motivated a wave of work on automated testbench generation~\cite{AutoBench2024,CorrectBench2025,UVM2_2025} and formal assertion synthesis~\cite{AssertLLM2024,Spec2Assertion2025,OrenesVera2023RTLFormal}; see~\cite{LLMVerifSurvey2026} for a comprehensive survey.

Verification is not a best-effort task. An undetected bug that escapes to silicon costs orders of magnitude more to fix than one caught before tape-out~\cite{Foster2015Trends}. Existing LLM-based approaches fall short of the quality bar this demands. Environment generation success rates remain low and reported coverage does not reflect specification requirements. No existing framework delivers end-to-end verification closure without human intervention, with coverage tied to specification behaviors.

The root cause is structural. Existing approaches~\cite{UVM2_2025,AutoBench2024,CorrectBench2025} generate each component through an independent single-turn LLM call with no shared context, so interface mismatches go undetected and cross-component errors cannot be resolved. Coverage models are built from RTL signal ranges rather than specification behaviors, so two designs with the same port list produce the same model regardless of their actual behavior. These two structural gaps give rise to three unresolved challenges.

\textbf{Challenge 1: Execution reliability.}
Assembling a verification environment needs iterative compilation, simulation, and repair. An agentic loop provides feedback closure, but agents without guardrails invoke the wrong tools, propagate poor artifacts across stages, and repeat incorrect repairs without progress~\cite{Yao2023ReAct,Wang2023AgentSurvey,Shinn2023Reflexion}. Prior work~\cite{UVM2_2025} does not address these failure modes. LLM reasoning and deterministic enforcement must therefore be architecturally separated, with each failure mode handled by a dedicated layer.

\textbf{Challenge 2: On-demand domain knowledge injection.}
No existing LLM-based verification framework provides a structured domain knowledge base. LLMs rely on general training knowledge alone, which is insufficient for complex modules. Injecting the full knowledge base inflates the context window, and injecting irrelevant knowledge actively misleads the model. Knowledge injection is therefore an on-demand dispatch problem. Each task requires different methodology knowledge and each design category requires different domain knowledge, both of which can be dispatched deterministically.

\textbf{Challenge 3: Specification-traceable coverage closure.}
Coverage models built from RTL signals measure stimulus breadth rather than specification compliance. When gaps remain, supplement strategies add stimulus without diagnosing why bins were missed, so the loop stalls without convergence. Tying every coverage bin to a named specification behavior solves both problems. Each gap then has a root cause and a targeted remedy, making convergence achievable. As shown in Fig.~\ref{fig:teaser}, prior work~\cite{AutoBench2024,CorrectBench2025} reaches at most L1 with code coverage only, while UVM$^2$~\cite{UVM2_2025} advances to L2 with signal-range functional coverage. \texttt{GoGoTB} targets L3--L4 by grounding coverage bins in specification behaviors.

We present \texttt{GoGoTB}, an agentic framework that achieves end-to-end verification closure with coverage tied to specification behaviors. Given RTL source, a natural-language specification, and an optional reference model, \texttt{GoGoTB} autonomously constructs and refines a verification environment until coverage targets are met. We summarize our key contributions as follows.

\begin{itemize}[leftmargin=*,topsep=2pt,itemsep=1pt]
\item We develop a three-layer execution control architecture that separates deterministic enforcement from LLM reasoning at tool-call, stage, and failure-recovery boundaries, achieving 100\% environment generation success across 8 designs and 7 backbone LLMs (Section~\ref{sec:harness}).

\item We design a two-tier knowledge system that dispatches methodology modules by deterministic stage-level mapping and historical environments by similarity gating, injecting domain knowledge on demand while suppressing irrelevant context (Section~\ref{sec:knowledge}).

\item We propose a specification-grounded coverage closure framework that derives testpoints across seven behavioral dimensions, classifies each residual gap into a seven-category root-cause taxonomy, and drives targeted remediation with a bounded stall-aware loop (Section~\ref{sec:closure}).

\item The complete \texttt{GoGoTB} framework and its evaluation benchmark of 8 open-source RTL designs will be open-sourced at a later date.
\end{itemize}

The rest of this paper is organized as follows. Section~\ref{sec:preliminary} introduces preliminaries. Section~\ref{sec:methodology} presents the methodology. Section~\ref{sec:exp} reports experimental results. Section~\ref{sec:conclusion} concludes the paper.

\begin{figure*}[!t]
  \centering
  \includegraphics[width=\textwidth]{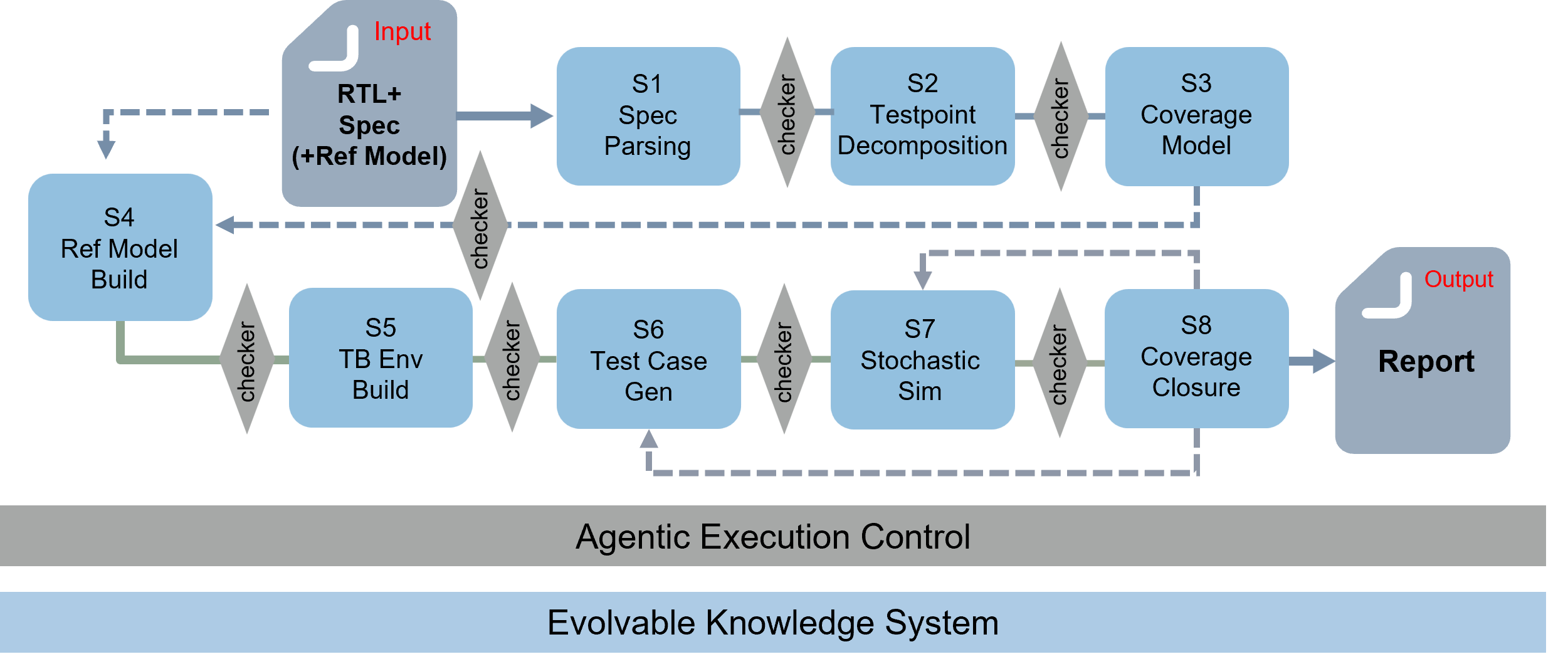}
  \caption{\texttt{GoGoTB} is an eight-stage agentic framework built on three subsystems. It includes agentic execution control, an evolvable knowledge system, and specification-grounded coverage closure.}
  \label{fig:overview}
\end{figure*}

\section{Preliminaries}
\label{sec:preliminary}

This section introduces the concepts underlying \texttt{GoGoTB} and surfaces the observations that motivate each architectural choice in Section~\ref{sec:methodology}.

\subsection{Digital IC Functional Verification}
\label{sec:prelim-dv}

\textbf{IC design flow and verification.}
Functional verification confirms that an RTL implementation satisfies its specification before synthesis sign-off. Because an undetected bug at this stage propagates into silicon and costs orders of magnitude more to fix after fabrication~\cite{Foster2015Trends}, verification constitutes the dominant fraction of front-end engineering effort~\cite{IEEEUVM2017,Foster2015Trends}. A complete UVM environment comprises a driver, monitor, scoreboard, functional coverage model, and reference model~\cite{IEEEUVM2017}, all traditionally hand-authored~\cite{Bergeron2003Writing} and consuming weeks per design.

\textbf{Testpoint decomposition.}
A verification campaign begins with \emph{testpoint decomposition}, enumerating every verifiable behavior in the specification~\cite{Wile2005FunctionalVerification}. A testpoint is a tuple $\langle s, r, c, d \rangle$ where $s$ is the stimulus, $r$ is the expected response, $c$ is the check method, and $d$ is the verification dimension. Testpoint completeness therefore sets a hard ceiling on verification quality. Any behavior not enumerated at this stage cannot be verified downstream.

\textbf{Coverage metrics.}
Verification coverage falls into two categories. \emph{Code coverage} tracks structural exercise through line, branch, and toggle metrics. \emph{Functional coverage}~\cite{Piziali2004Coverage} measures which specification-defined scenarios have been exercised using coverpoints and bins.

Coverage model quality is determined at authoring time, not at measurement time. A model built from RTL signal ranges can reach 100\% bin coverage while leaving critical specification behaviors unverified. Two designs with the same port list produce identical models despite doing entirely different things. Coverage must therefore be grounded in the specification before simulation begins. Once every bin is anchored to a specification behavior, gaps become diagnosable and closure becomes principled.

\textbf{Simulation stack.}
The dominant commercial stack pairs Synopsys VCS~\cite{TOOL_vcs} or Cadence Xcelium~\cite{Xcelium2023} with SystemVerilog/UVM testbenches. Prior LLM-based approaches~\cite{UVM2_2025} follow the same path and generate SystemVerilog components directly. LLMs produce substantially lower first-pass correct code in SystemVerilog than in Python, making Python-based cocotb~\cite{cocotb2023} a better fit for LLM-driven generation. Verilator~\cite{snyder2018verilator} emits structured, machine-parseable diagnostics that enable deterministic error classification without LLM involvement.

\subsection{From Language Models to Agentic Systems}
\label{sec:prelim-agent}

\textbf{LLMs and their structural limitation.}
An LLM autoregressively samples an output sequence given a fixed prompt:
\begin{equation}
  P(y_{1:m}\mid x_{1:n}) = \prod_{t=1}^{m} P(y_t \mid x_{1:n}, y_{1:t-1}).
\end{equation}
Here $x_{1:n}$ is the input prompt, $y_{1:m}$ is the generated output sequence, and each token $y_t$ is sampled conditioned on the prompt and all previously generated tokens.
This is a single stateless mapping with no memory across turns and no mechanism to act on the world. All prior LLM-based verification works~\cite{UVM2_2025,UVLLM2024,AutoBench2024,CorrectBench2025,LLM4DV2023,LLMVerifSurvey2026} operate within this paradigm, which is structurally insufficient for verification environment construction.

\textbf{The LLM agent abstraction.}
An agent augments an LLM with persistent state, tool access, and a reasoning-action-observation loop~\cite{Yao2023ReAct,Wang2023AgentSurvey}. At each step the agent produces an action conditioned on accumulated context and updates its state from the observed result. This feedback closure enables the iterative repair and failure-adaptive recovery that verification construction requires.

The agent abstraction alone is insufficient. Unconstrained agentic loops corrupt protected design files, disable checking logic, and loop indefinitely without progress. These are control failures from the absence of an enforcement layer. LLM reasoning and deterministic enforcement must therefore be architecturally separated so that reliability depends on the control architecture rather than on any particular model.

\textbf{Knowledge injection and the case for a domain knowledge base.}
Industrial verification demands two types of expertise. The first is methodology knowledge covering how to structure testbenches, apply constrained-random patterns, analyze RTL, and close coverage. The second is design-specific knowledge covering protocol timing, FSM sequencing, cryptographic patterns, and simulator constraints for a particular design category. Neither type is reliably captured in LLM training data and no existing LLM-based verification framework provides a structured domain knowledge base.

Injecting the full knowledge base at every stage inflates the context window and dilutes the relevant signal, so injection must be on demand. These two types of knowledge also require different dispatch mechanisms. Methodology knowledge is predictable from the pipeline stage definition and can be dispatched deterministically. Design-specific knowledge depends on how closely the current design matches past ones. A low-similarity reference causes the LLM to apply the wrong template with high confidence, which is worse than injecting nothing. Design-specific knowledge must therefore be controlled by similarity gating and suppressed when confidence is insufficient.

\section{Methodology}
\label{sec:methodology}

\subsection{\texttt{GoGoTB} Overview}
\label{sec:overview}

\texttt{GoGoTB} takes RTL source, a natural-language specification, and a reference model as input.
Fig.~\ref{fig:overview} illustrates the eight-stage pipeline, which runs in four phases.
\emph{Specification Analysis} covers S1--S2 and parses the specification into a design profile and derives a testpoint inventory across seven behavioral dimensions.
\emph{Environment Construction} covers S3--S5 and builds the coverage model, reference model, and testbench from those testpoints.
\emph{Stochastic Simulation} covers S6--S7 and runs test stimuli under an adaptive seed policy and accumulates code and functional coverage.
\emph{Coverage Closure} covers S8 and diagnoses residual gaps by root cause and drives targeted remediation until closure criteria are met.
Each stage runs as an agentic node. Deterministic quality checks at each stage boundary block broken outputs before they reach the next stage.

Reliable closure across this pipeline requires three subsystems, each realizing one design insight from Section~\ref{sec:intro}.
\begin{enumerate}[leftmargin=*,topsep=2pt,itemsep=1pt]
  \item An \emph{agentic execution control layer} (Section~\ref{sec:harness}) addresses three independent failure modes through three dedicated layers. Each layer targets one failure mode: tool-call boundaries, stage boundaries, and repeated repair attempts.
  \item An \emph{evolvable knowledge system} (Section~\ref{sec:knowledge}) injects the right knowledge at the right stage. A Skill Library dispatches methodology modules by stage and design category. A Reference Library selects past environments by structural similarity and suppresses dispatch when no close match exists.
  \item \emph{Specification-grounded coverage closure} (Section~\ref{sec:closure}) anchors every coverage bin to a named specification behavior before simulation begins. When gaps remain, each uncovered bin has an unambiguous root cause and a targeted remedy drawn from a seven-category taxonomy.
\end{enumerate}

The execution control layer wraps all eight stages. The knowledge system injects domain knowledge at the entry point of each stage. Coverage closure runs across stages S1--S3 and S6--S8.

\subsection{Agentic Execution Control}
\label{sec:harness}

\begin{figure}[t]
  \centering
  \includegraphics[width=\columnwidth]{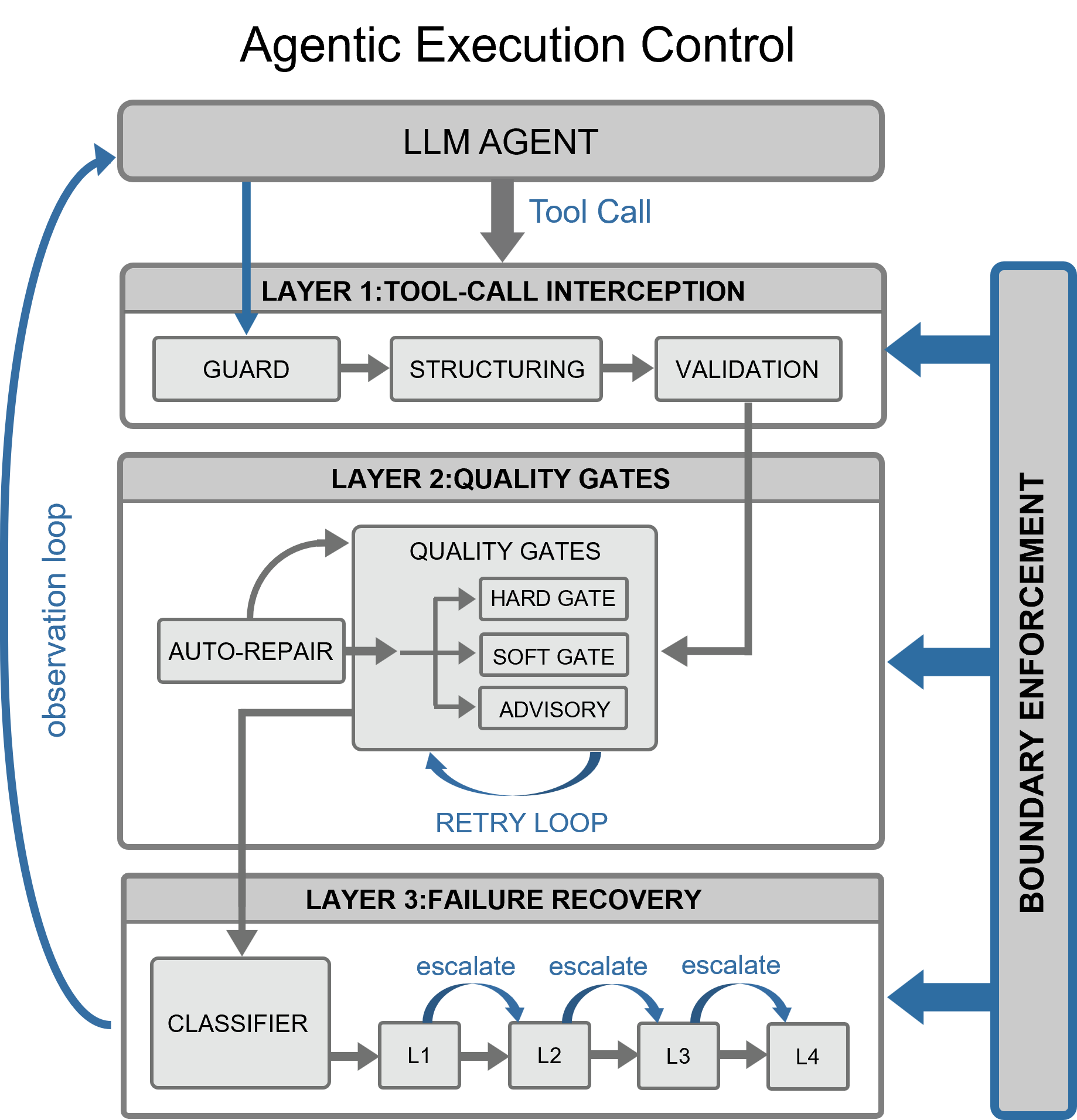}
  \caption{Three-layer agentic execution control. Layer~1 intercepts every tool call with guard, structuring, and validation steps. Layer~2 enforces quality gates at stage boundaries with deterministic auto-repair. Layer~3 escalates repair scope through four levels when earlier fixes fail. Boundary enforcement applies as a cross-cutting constraint across all layers.}
  \label{fig:harness}
\end{figure}

Reliable loop execution requires that every LLM action passes through a deterministic enforcement boundary. \texttt{GoGoTB} introduces three dedicated layers for this (Fig.~\ref{fig:harness}). Layer~1 controls every tool call, output, and session boundary so the LLM always reasons from clean, structured information. Layer~2 checks every stage boundary and blocks broken outputs before they reach the next stage. Layer~3 classifies each failure pattern and escalates repair scope rather than repeating the same action.

\textbf{Layer 1: Tool-Call Interception.}
Without an intermediary between the LLM and its tools, the LLM invokes the wrong tools, reads raw output directly, and decides when a session is done. All three contact points are sources of unreliable behavior. Every tool call therefore passes through three steps in sequence (Fig.~\ref{fig:harness}). The \textbf{Guard} step checks that the LLM is invoking the right tool for the current stage and blocks incorrect calls before execution. The \textbf{Structuring} step converts raw compiler or simulator output into a clean structured diagnostic so the LLM reasons from root causes rather than surface noise. The \textbf{Validation} step verifies at the session boundary that the session produced complete and compilable artifacts before it closes. If the same command fails three times in a row with no change, a redirect prompt is injected to break the stuck loop. The structured result is returned to the LLM through the observation loop. All three steps run independently of LLM decisions.

\textbf{Layer 2: Quality Gates.}
Poor artifacts from one stage propagate silently to the next, causing failures that are attributed to the wrong stage and making diagnosis impossible. Layer~2 first runs an \textbf{Auto-Repair} pass that fixes common structural errors without LLM involvement. The output then passes through three gates in order of strictness. The \textbf{Hard Gate} checks for compilation success and mandatory artifacts and always blocks on failure. The \textbf{Soft Gate} checks for testpoint completeness and signal-name consistency. It blocks on failure in early attempts and softens to a warning as attempt count rises. The \textbf{Advisory Gate} performs a semantic review without blocking progress. On failure, the output is returned to the LLM through the retry loop and re-enters the gates on the next attempt.

\textbf{Layer 3: Failure Recovery.}
When a repair attempt fails, repeating the same action without changing scope makes no progress and exhausts the retry budget. Layer~3 first passes the failure through a classifier that identifies the error category and injects targeted feedback into the next attempt. Each time the attempt count crosses a threshold, the repair automatically escalates to the next level along the L1--L4 chain. \textbf{L1} applies a minimal patch to a single file. \textbf{L2} reads full cross-file context before editing. \textbf{L3} refactors the problematic component entirely. \textbf{L4} falls back to a minimal-correct implementation. Each escalation step expands the repair scope so the loop converges rather than stalls.

\textbf{Boundary Enforcement.}
Boundary Enforcement applies across all three layers as a cross-cutting constraint (Fig.~\ref{fig:harness}). Each stage can only access the artifacts it needs. The RTL source and the reference model are immutable to all later stages. The reference model is derived from the specification alone and has no visibility into RTL implementation details. This isolation guarantee ensures that any simulation mismatch cannot originate from a corrupted oracle and therefore points to a real design defect.

\subsection{Evolvable Knowledge System}
\label{sec:knowledge}

\begin{figure}[t]
  \centering
  \includegraphics[width=\columnwidth]{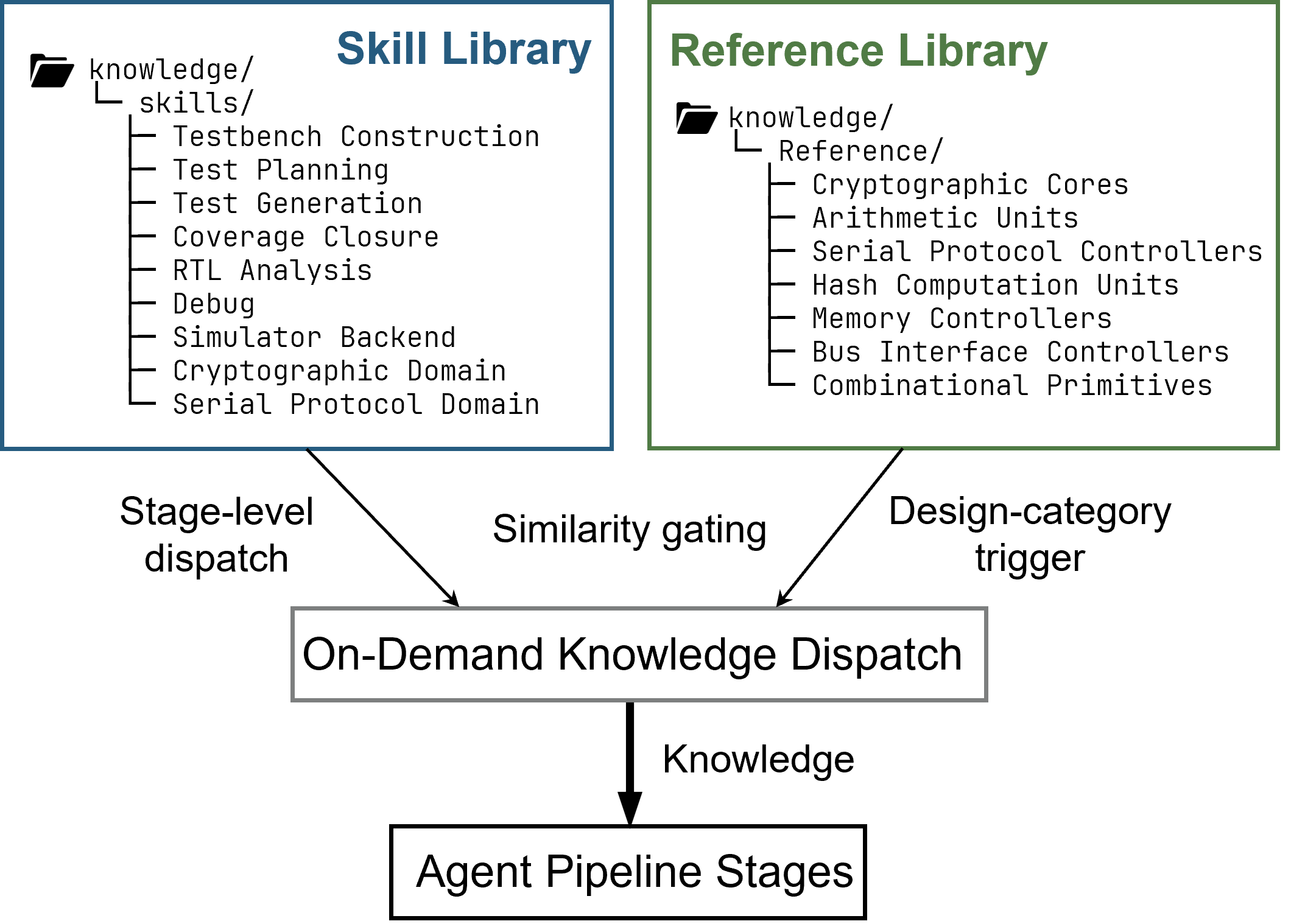}
  \caption{Two-tier evolvable knowledge system. The Skill Library dispatches methodology modules by a deterministic stage-level map. The Reference Library selects historical environments by similarity gating and suppresses dispatch when no close match exists.}
  \label{fig:knowledge}
\end{figure}

Knowledge injection is an on-demand dispatch problem. Stage-level methodology knowledge is predictable from the stage definition and can be dispatched deterministically. Design-specific knowledge depends on structural similarity to the current design and must be controlled by similarity gating. Both dispatch mechanisms require no LLM involvement and cannot silently apply the wrong knowledge. The two-tier architecture in Fig.~\ref{fig:knowledge} realizes these two mechanisms separately.

\textbf{Skill Library.}
The Skill Library delivers stage-level methodology knowledge through deterministic dispatch. It holds ten skill modules covering testbench construction, test planning, coverage closure, RTL analysis, simulator backend, cryptographic design patterns, serial protocol patterns, and more. Before each pipeline stage runs, a stage-level map selects which modules apply (Fig.~\ref{fig:knowledge}). Selection requires no LLM involvement. To keep context small, only a table of contents arrives at stage start and full module content loads on demand. Adding a new design category requires only adding a new file to the knowledge directory.

\textbf{Reference Library.}
The Reference Library delivers design-specific historical knowledge through similarity gating. A design-category trigger activates the library based on the design category detected at specification-parse time. Each entry holds the RTL, specification, testbench, and testpoints from a real verified design. When a new design arrives, the system scores it against the library across three structural levels. A close match dispatches the reference as a direct template. A partial match dispatches it as a style guide. A poor match suppresses dispatch entirely so the LLM receives nothing rather than a misleading reference. This suppression is the key design choice. The LLM only receives historical knowledge when that knowledge genuinely applies.

\subsection{Specification-Grounded Coverage Closure}
\label{sec:closure}

Coverage quality is determined at model authoring time, not at measurement time. Grounding every bin in a specification behavior before simulation makes gaps diagnosable and convergence achievable, because the reason a bin is uncovered must belong to a finite set of root causes. As listed in Table~\ref{tab:gap}, these causes include insufficient stimulus, missing multi-step sequences, cross-coverage constraint mismatches, unreached FSM states, timing-dependent conditions, edge transitions, and inadequate seed sampling. Each cause demands a different targeted remedy. The loop terminates when no addressable cause remains. Without specification grounding, none of these categories is diagnosable and the supplement loop runs without principled convergence.

\textbf{Testpoint decomposition (S1--S2).}
To anchor every bin to a named specification behavior before simulation begins, S1 parses the RTL and specification into a design profile. S2 derives testpoints across seven behavioral dimensions. \textbf{D1} covers data boundaries, \textbf{D2} control flow, \textbf{D3} timing constraints, \textbf{D4} FSM state transitions, \textbf{D5} protocol compliance, \textbf{D6} error injection, and \textbf{D7} microarchitectural interaction. These seven dimensions cover all verifiable behavior types in a design specification. The resulting testpoint set establishes the ceiling on verification completeness.

\textbf{Coverage model and simulation (S3, S6--S7).}
With testpoints in place, S3 converts each one into a coverage model element where every bin carries a named behavioral claim traceable to the specification. This transforms the coverage model from a signal-value partition into a set of behavioral assertions. S6 then generates test cases per testpoint group. S7 runs them under an adaptive seed policy and stops when consecutive batches yield no gain, marking the boundary of what stochastic simulation can reach on its own.

\textbf{Gap diagnosis and closure (S8).}
Because every bin is specification-grounded, each residual gap maps directly to a cause in Table~\ref{tab:gap}. S8 classifies each uncovered bin and directs it to the targeted remedy. Structurally unreachable bins are removed from the denominator. A stall detector terminates the loop when two consecutive rounds show no improvement. The sign-off report records evidence for every covered bin and a root-cause disposition for every residual gap. The verification outcome is fully auditable with no gap silently dropped.

\begin{table}[t]
\centering
\caption{Gap root-cause taxonomy and remediation routing in S8.}
\label{tab:gap}
\footnotesize
\setlength{\tabcolsep}{4pt}
\renewcommand{\arraystretch}{1.2}
\begin{tabular}{ll}
\toprule
\textbf{Root-cause category} & \textbf{Targeted remedy} \\
\midrule
\textsc{Stimulus Gap}    & Directed test generation \\
\textsc{Sequence Gap}    & Multi-transaction directed tests \\
\textsc{Cross Hole}      & Constraint refinement \\
\textsc{State Unreached} & State-routing sequences \\
\textsc{Timing Dependent}& Cycle-precise stimulus \\
\textsc{Transition Gap}  & Edge-condition tests \\
\textsc{Seed Insufficient}& Additional seed expansion \\
\bottomrule
\end{tabular}
\end{table}

\section{Experimental Evaluation}
\label{sec:exp}

\smallskip
\noindent\textbf{RQ1} (Sec.~\ref{subsec:rq1}): Does the three-layer execution control architecture achieve reliable environment generation across diverse designs and LLM backends?

\smallskip
\noindent\textbf{RQ2} (Sec.~\ref{subsec:rq2}): Does specification-grounded testpoint decomposition produce a richer coverage model and higher functional coverage than signal-space generation?

\smallskip
\noindent\textbf{CS1} (Sec.~\ref{subsec:casestudy1}): Does domain knowledge injection resolve total functional coverage failure caused by a knowledge gap rather than a model capability gap?

\smallskip
\noindent\textbf{CS2} (Sec.~\ref{subsec:casestudy2b}): Does specification-grounded closure converge through root-cause diagnosis across all seven gap categories?

\subsection{Experimental Setup}
\label{subsec:setup}

\textbf{Platform.}
Intel Xeon Gold 6248R with 48 cores, 512\,GB DDR4, and Ubuntu~22.04 LTS.
Verilator~5.024~\cite{snyder2018verilator} with cocotb~2.0.1~\cite{cocotb2023}.
\texttt{GoGoTB} is implemented on the CodeBuddy Agent SDK~\cite{codebuddy2025}.
RQ1 evaluates seven backbone LLMs: Kimi-K2.5, GLM-5.1, Minimax-M2.7, Claude Opus~4.6, Claude Sonnet~4.6, GPT-5.4, and DeepSeek-V3.2.
RQ2 and all case studies use Kimi-K2.5 as the primary model.

\textbf{Benchmark suite.}
Table~\ref{tab:benchmark} lists eight benchmark designs ranging from 291 to 1{,}424 lines and 1 to 9 modules, spanning combinational datapaths, pipelined cryptographic cores, serial protocols, and stateful memory controllers.

\begin{table*}[!t]
\centering
\setlength{\tabcolsep}{6pt}
\renewcommand{\arraystretch}{1.4}
\caption{Benchmark designs used for evaluation ($^\dagger$\,=\,minor Verilator patch).}
\label{tab:benchmark}
\footnotesize
\begin{tabular*}{\textwidth}{@{}m{1.1cm}m{9.5cm}p{3.2cm}@{\extracolsep{\fill}}cc@{}}
\toprule
\multirow{2}{*}{\textbf{Design}} & \multirow{2}{*}{\textbf{Description}} & \textbf{Source} & \textbf{Module} & \textbf{Line} \\
 & & & \textbf{Counts} & \textbf{Counts} \\
\midrule
AES$^\dagger$    & Encrypts a 128-bit plaintext with a 128-bit key using the AES algorithm, outputting 128-bit ciphertext. & UVM$^2$~\cite{UVM2_2025} &  8 &  684 \\ \hline
ALU$^\dagger$    & A 32-bit unit performing IEEE\,754 FP arithmetic, logical operations (OR, AND, XOR), shifts, and FP-to-integer conversion. & UVM$^2$~\cite{UVM2_2025} &  7 &  409 \\ \hline
UART   & Facilitates serial communication between a host and peripherals, supporting configurable baud rates and stop bits. & \cite{timrudy2020uart} &  4 &  387 \\ \hline
SPI    & A lightweight SPI controller supporting Master mode with FIFO-based data transfer and interrupt-driven operation. & This work &  3 &  291 \\ \hline
I2C    & An I2C slave controller handling address detection, data transfer, clock stretching, and ACK/NACK signaling. & \cite{forencich2017i2c} &  3 &  470 \\ \hline
SHA256 & A cryptographic unit that computes a 256-bit hash value from an input message using the SHA-256 algorithm. & \cite{strombergson2013sha256} &  4 & \textbf{1,142} \\ \hline
SM4    & A synchronous SM4 encryption/decryption core supporting both modes via a 128-bit key and data interface. & \cite{gongxunwu2020sm4} &  9 & \textbf{1,424} \\ \hline
SDRAM  & A synchronous DRAM controller managing ACTIVATE, READ, WRITE, and PRECHARGE commands with timing enforcement. & \cite{terasic2012de0nano} &  1 &  433 \\
\bottomrule
\end{tabular*}
\end{table*}

AES and ALU are taken from the UVM$^2$ benchmark suite, the only two designs that UVM$^2$ makes publicly available. The remaining six designs were selected from high-quality open-source RTL repositories to cover a broad range of design categories not represented in UVM$^2$. The complete benchmark package, including specifications, verification environments, and testpoints for all eight designs, will be open-sourced at a later date. All experiments run fully automated without human intervention and provide a reference model as input.

\textbf{Metrics.}
\emph{SRG} measures whole-environment success requiring zero-error compilation, deadlock-free simulation, and non-zero coverage. For design $d_i$ under model $\mathcal{M}$:
\begin{equation}
  \mathrm{SRG}(\mathcal{M}) = \frac{1}{N}\sum_{i=1}^{N} \mathbf{1}[\,\text{compile} \wedge \text{simulate} \wedge \text{cov}>0\,].
\end{equation}
\emph{Code coverage} reports per-design line, branch, and toggle ratios under Kimi-K2.5.
\emph{Functional coverage} reports the fraction of specification-derived bins hit after stochastic simulation.
\emph{Coverage model richness} reports coverpoint and cross-coverage counts per design, compared against UVM$^2$.

\subsection{RQ1: Environment Generation Reliability}
\label{subsec:rq1}

\begin{figure}[t]
  \centering
  \includegraphics[width=\columnwidth]{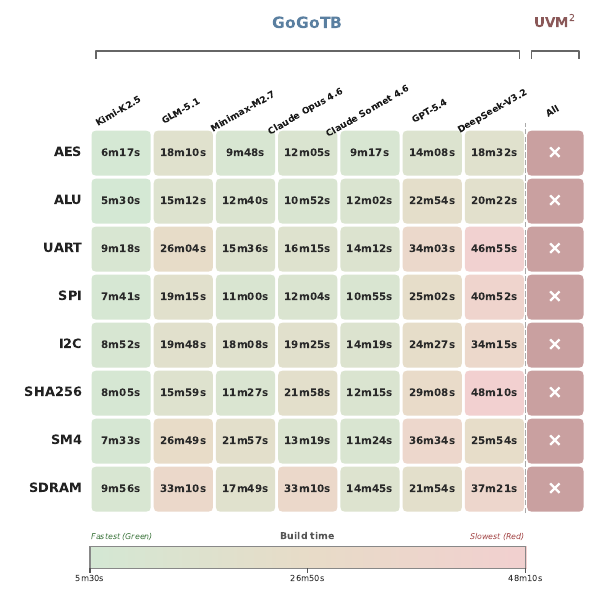}
  \caption{Environment generation success rate and build time across designs and backbone LLMs.}
  \label{fig:srg_heatmap}
\end{figure}

Fig.~\ref{fig:srg_heatmap} reports SRG and wall-clock build time for both systems.
\texttt{GoGoTB} achieves \textbf{100\% SRG} on 8 of 8 designs across all seven backbone LLMs. Build times range from 5m30s to 48m10s. The result holds across model families, parameter scales, and training corpora. The enforcement layers drive convergence, not the backbone LLM.

SRG measures whole-environment success. A high per-component generation rate does not imply that components assemble into a runnable environment, because components generated in isolation diverge in interface assumptions and the assembled environment fails even when each component passes its own syntax check.

UVM$^2$ achieves \textbf{0\% SRG} on all eight designs. We cloned the UVM$^2$~\cite{UVM2_2025} repository and ran each design four to six times. UVM$^2$ generates each component through a single LLM call with no shared context and no generate-debug-iterate loop. Compilation failures and simulation deadlocks accumulate without any recovery mechanism. These failures are architectural and follow directly from the absence of enforcement and iteration, not from the capability of any specific LLM.

The 100\% vs.\ 0\% contrast across seven LLMs confirms that the three-layer enforcement architecture drives loop convergence regardless of model choice.

\subsection{RQ2: Coverage Achievement and Model Quality}
\label{subsec:rq2}

Table~\ref{tab:coverage} presents results under Kimi-K2.5. UVM$^2$ is the only prior system that attempts functional coverage, making it the only valid baseline. \texttt{GoGoTB} derives every coverpoint from a named specification behavior. Each bin tests whether a specific design intent was exercised. UVM$^2$ partitions transaction signal ranges into bins with no connection to specification meaning.

\definecolor{rowgray}{gray}{0.92}

\begin{table*}[t!]
\centering
\caption{RQ2: cross-system coverage results under Kimi-K2.5 ($\times$\,=\,compile/sim failed; {---}\,=\,not available; $^\dagger$\,UVM$^2$ SRG 0/8; UVM$^2$ subscriber open-sourced only for AES and ALU).}
\label{tab:coverage}
\footnotesize
\setlength{\tabcolsep}{5pt}
\renewcommand{\arraystretch}{1.2}
\begin{tabular*}{\textwidth}{@{\extracolsep{\fill}}llcccccccc c@{}}
\toprule
\multirow{2}{*}{\textbf{DUT}} & \multirow{2}{*}{\textbf{Cat.}} &
  \multicolumn{6}{c}{\textbf{\texttt{GoGoTB}}} &
  \multicolumn{3}{c}{\textbf{UVM$^2$}$^\dagger$} \\
\cmidrule(lr){3-8}\cmidrule(l){9-11}
& &
  \textbf{Coverpoint} & \textbf{Cross} &
  \textbf{Line} & \textbf{Branch} & \textbf{Toggle} & \textbf{Func} &
  \textbf{Coverpoint} & \textbf{Cross} & \textbf{L/Br/T/F} \\
\midrule
  AES    & CC  &  6 &  2 &  99.3 &  93.7 &  93.7 &  84.2 &  3 &  1 & $\times$ \\
  ALU    & CS  &  7 &  4 &  92.2 &  96.9 &  96.8 &  95.6 &  3 &  0 & $\times$ \\
  UART   & SP  & 12 &  2 &  98.7 & 100.0 & 100.0 &  75.3 & {---} & {---} & $\times$ \\
  SPI    & SP  & 11 &  3 & 100.0 & 100.0 & 100.0 &  93.3 & {---} & {---} & $\times$ \\
  I2C    & SP  & 14 &  3 &  99.1 &  95.2 &  94.7 &  84.1 & {---} & {---} & $\times$ \\
  SHA256 & SC  & 10 &  2 &  99.8 &  99.9 & 100.0 &  89.9 & {---} & {---} & $\times$ \\
  SM4    & SC  & 20 &  7 &  98.9 & 100.0 &  99.4 &  72.3 & {---} & {---} & $\times$ \\
  SDRAM  & MC  &  9 &  3 &  99.0 &  91.9 &  91.2 &  85.3 & {---} & {---} & $\times$ \\
\midrule
  \textbf{Avg.} & &
  \textbf{10.8} & \textbf{3.0} &
  \textbf{98.4} & \textbf{97.2} & \textbf{97.0} & \textbf{83.2} &
  --- & --- & \textbf{0/8} \\
\bottomrule
\end{tabular*}
\par\vspace{2pt}
{\footnotesize Cat.: CC\,=\,Comb.\ crypto; CS\,=\,Comb./seq.; SP\,=\,Serial proto.; SC\,=\,Sequential crypto; MC\,=\,Memory ctrl.
  {---}\,=\,UVM$^2$ subscriber not open-sourced for this design.}
\end{table*}

\texttt{GoGoTB} averages \textbf{98.4\%} line, \textbf{97.2\%} branch, \textbf{97.0\%} toggle, and \textbf{83.2\%} functional coverage. Five designs reach ${\geq}99\%$ line coverage and all exceed 91\% branch coverage.

UVM$^2$ achieves 0\% SRG, so no functional coverage numbers are available. We compare coverage model quality on AES and ALU, the only two designs UVM$^2$ open-sources. UVM$^2$ produces 3 coverpoints and 1 cross on AES. \texttt{GoGoTB} produces 6 coverpoints and 2 crosses. Each \texttt{GoGoTB} coverpoint maps to a named specification behavior such as key shape, NIST anchor hit, and plaintext change. Each UVM$^2$ bin is a numeric interval on a raw signal with no behavioral claim. For the remaining six designs UVM$^2$ provides no coverage model at all.

Code coverage exceeds 97\% while functional coverage averages 83.2\%. This gap reflects the inherent difficulty of specification-level verification. Reaching a code line is easier than hitting the precise multi-signal condition that a specification-derived bin requires. Closing this gap requires advances in LLM-driven directed test generation.

\subsection{CS1: Knowledge-Driven Environment Construction on UART}
\label{subsec:casestudy1}

\begin{figure}[!t]
  \centering
  \includegraphics[width=\columnwidth]{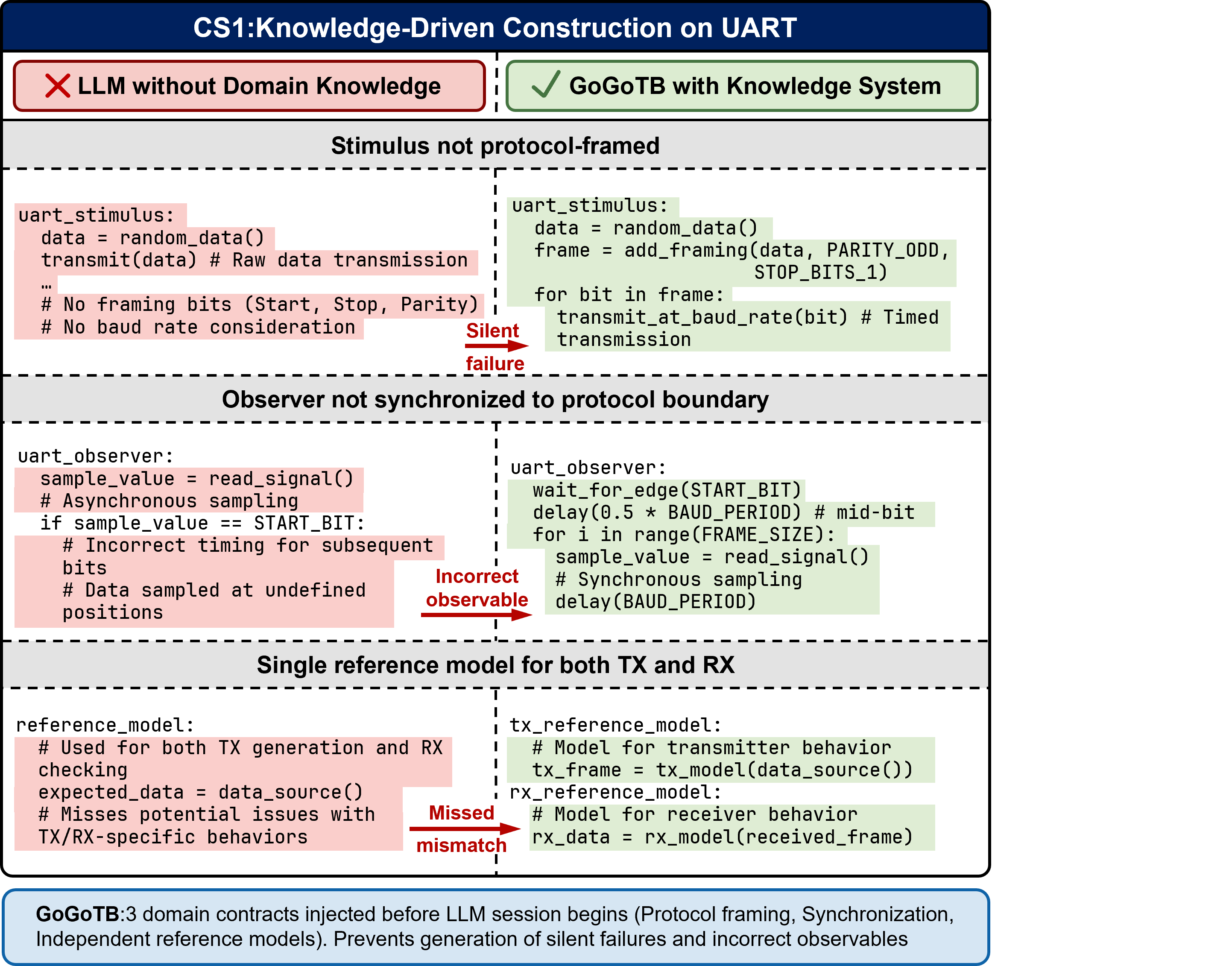}
  \caption{CS1: Three silent failures (left) caused by missing domain knowledge and the corresponding fixes (right) from \texttt{GoGoTB} domain contracts. Each row addresses one failure: protocol framing (top), observer synchronization (middle), and reference model separation (bottom).}
  \label{fig:cs2_knowledge}
\end{figure}

CS1 examines whether domain knowledge injection resolves coverage failures caused by a knowledge gap rather than a model capability gap. We select UART because its protocol framing, baud-rate timing, and direction-specific TX/RX behavior are domain conventions that no general-purpose LLM reliably encodes.

We first run \texttt{GoGoTB} on UART with the knowledge system disabled. All other subsystems remain active. Under this configuration, the LLM agent generates a testbench with three silent failures shown in Fig.~\ref{fig:cs2_knowledge}. The stimulus drives raw signal transitions without protocol framing, so the DUT never sees a valid UART frame. The observer samples asynchronously instead of synchronizing to the START\_BIT boundary, so it produces bit-shifted data that never matches the expected output. A single reference model serves both TX and RX paths and generates wrong expectations even when the DUT responds correctly. These three failures form a closed chain from stimulus to oracle. The environment compiles and simulates without error, yet the coverage report shows all functional bins at \textbf{0\%}.

\texttt{GoGoTB} injects three domain contracts before the LLM session to eliminate these failures. The protocol-framing contract enforces correct UART frame construction with start bit, parity, and stop bits. The synchronization contract aligns the observer to the mid-bit sampling point at $0.5 \times \text{BAUD\_PERIOD}$. The independent-models contract instructs the LLM agent to model TX and RX behavior separately so that it captures direction-specific expectations. Functional coverage rises from 0\% to \textbf{75.3\%}. The only variable changed is the knowledge system, confirming that the failure was a knowledge gap, not a model capability gap. Functional coverage does not reach 100\% because the LLM agent still struggles to generate directed tests for protocol-specific corner cases such as multi-frame error recovery and baud-rate edge conditions.

\subsection{CS2: Coverage-Driven Closure on I2C}
\label{subsec:casestudy2b}

CS2 examines whether specification-grounded closure remains principled and convergent across all seven root-cause categories. We select I2C because its stateful protocol behavior exercises all seven categories simultaneously.

After the initial test suite completes regression, functional coverage reaches \textbf{56.0\%} across 132 specification-derived bins. The system then classifies every uncovered bin by root cause, routes each to its targeted remedy, and generates new tests. This first supplement round advances coverage to \textbf{71.6\%}. The loop repeats the same root-cause analysis and targeted generation over multiple iterations until coverage converges at \textbf{84.1\%}, a total gain of \textbf{+28\,pp}. Each decision follows from the root-cause label on the uncovered bin, not from a generic retry strategy.

After the loop terminates, the system analyzes the 24 residual bins. Stimulus Gap and Sequence Gap account for 16, Cross Hole and State Unreached account for 5, and the remaining 3 are Timing Dependent, Transition Gap, and Seed Insufficient. Stimulus and sequence gaps mean that the system identified the correct bin and knew what stimulus was needed, but the LLM agent could not produce a test that hits the required condition. The coverage model itself is not the bottleneck. The bottleneck is the ability of the LLM agent to generate precise directed tests for complex multi-step protocol scenarios.

\begin{figure}[!t]
  \centering
  \includegraphics[width=\columnwidth]{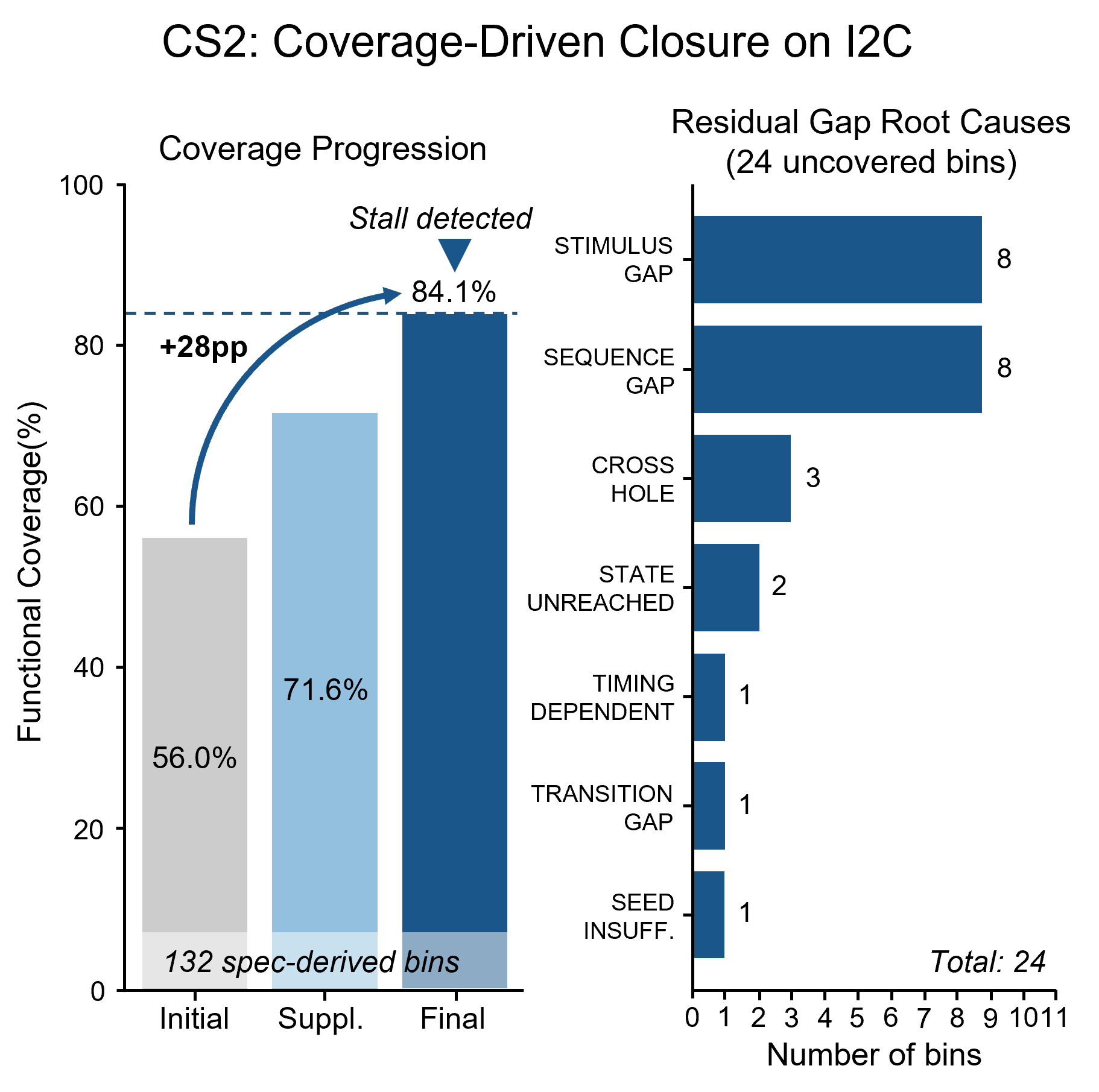}
  \caption{CS2: I2C coverage progression and root-cause distribution of 24 residual gaps. Stimulus and sequence gaps dominate, indicating that the system correctly identifies what to test but the LLM agent cannot yet produce the required directed tests.}
  \label{fig:cs3_coverage}
\end{figure}

\section{Conclusion}
\label{sec:conclusion}

\texttt{GoGoTB} demonstrates that fully automated IC functional verification with coverage tied to specification behaviors is achievable when a single system co-designs an agentic execution control layer, an evolvable knowledge system, and specification-grounded coverage closure.

The three-layer execution control makes environment generation succeed across all 8 designs and 7 backbone LLMs. The evolvable knowledge system eliminates domain knowledge failures that would otherwise cause silent coverage collapse, as demonstrated by the 0\% to 75.3\% functional coverage recovery on UART. Specification-grounded coverage closure drives functional coverage to an average of 83.2\% across all designs, with a +28\,pp gain through iterative root-cause diagnosis on I2C alone.

The primary limitation is directed test generation for complex multi-step protocol scenarios. Stimulus Gap and Sequence Gap account for the majority of residual bins across designs. Improving LLM-driven directed test generation and integrating formal methods to cover structurally hard-to-reach states are the most impactful directions for future work.

On the current benchmark, \texttt{GoGoTB} averages 98.4\% line, 97.2\% branch, 97.0\% toggle, and 83.2\% functional coverage with fully auditable sign-off reports.

\clearpage
{
\small

}

\end{document}